\documentclass[journal]{IEEEtran}

\IEEEoverridecommandlockouts
\usepackage{cite}

\ifCLASSINFOpdf
\else
\fi
\usepackage{graphicx}
\usepackage[]{footmisc}

\usepackage{amsmath}
\usepackage{amsfonts}
\usepackage{tikz}
\usepackage{textcomp}
\usepackage{hyperref}
\usepackage{lipsum}

\newcommand\copyrighttext{%
	\footnotesize \textcopyright © 2017 IEEE. Personal use of this material is permitted. Permission from IEEE must be obtained for all other uses, in any current or future media, including reprinting/republishing this material for advertising or promotional purposes, creating new collective works, for resale or redistribution to servers or lists, or reuse of any copyrighted component of this work in other works.
}

\newcommand\copyrightnotice{%
	\begin{tikzpicture}[remember picture,overlay]
	\node[anchor=south,yshift=3pt] at (current page.south) {\fbox{\parbox{\dimexpr\textwidth-\fboxsep-\fboxrule\relax}{\copyrighttext}}};
	\end{tikzpicture}%
}

\begin{document}
%
\title{Neural networks catching up with finite differences in solving partial differential equations in higher dimensions}
%
%
%

\author{V.I. Avrutskiy
\thanks{V.I. Avrutskiy is with the Department of Aeromechanics and Flight Engineering of Moscow Institute of Physics and Technology, Institutsky lane 9, Dolgoprudny, Moscow region, 141700, e-mail: avrutsky@phystech.edu}
}

%
%

\markboth{Submitted to IEEE Transactions on Neural Networks and Learning Systems}%
{Shell \MakeLowercase{\textit{et al.}}: Bare Demo of IEEEtran.cls for IEEE Journals}
%



\maketitle
\copyrightnotice
\begin{abstract}
Fully connected multilayer perceptrons are used for obtaining numerical solutions of partial differential equations in various dimensions. Independent variables are fed into the input layer, and the output is considered as solution's value. To train such a network one can use square of equation's residual as a cost function and minimize it with respect to weights by gradient descent. Following previously developed method, derivatives of the equation's residual along random directions in space of independent variables are also added to cost function. Similar procedure is known to produce nearly machine precision results using less than 8 grid points per dimension for 2D case. The same effect is observed here for higher dimensions: solutions are obtained on low density grids, but maintain their precision in the entire region. Boundary value problems for linear and nonlinear Poisson equations are solved inside 2, 3, 4, and 5 dimensional balls. Grids for linear cases have 40, 159, 512 and 1536 points and for nonlinear 64, 350, 1536 and 6528 points respectively. In all cases maximum error is less than $8.8\cdot10^{-6}$, and median error is less than $2.4\cdot10^{-6}$. Very weak grid requirements enable neural networks to obtain solution of 5D linear problem within 22 minutes, whereas projected solving time for finite differences on the same hardware is 50 minutes. Method is applied to second order equation, but requires little to none modifications to solve systems or higher order PDEs.
\end{abstract}


%
\IEEEpeerreviewmaketitle

\section{Introduction}
\IEEEPARstart{P}{artial} differential equations have an enormous number of applications and are fundamental for predictions of many natural phenomena. Sometimes pencil and paper are enough to obtain their solutions as closed-form expressions or infinite series but in most cases numerical methods are the only remedy. A vast number of those were developed\cite{smith1985numerical,chorin1968numerical,eymard1999finite,bassi1997high,taflove2005computational,hairer2006geometric}. Operating with fixed resources, they all have to use some finite way to describe functions they are trying to find. Either by values on a grid or by set of simple functions defined inside minute volumes, numerical methods are looking for a suitable finite set of real, or rather, rational parameters that can be used to construct a solution. This paper describes a solving method that represents functions using neural networks. Here is an example of how one can look like:
\[
u(x)=\sum_{k}W_{k}^{3}\sigma\left(\sum_{j}W_{kj}^{2}\sigma\left(\sum_{i}W_{ji}^{1}\sigma\left(W_{i}^{0}\cdot x\right)\right)\right)
\]
This particular expression represents a real valued function $u$ of one argument $x$. Here $\sigma$ is a special kind of nonlinear scalar mapping that is applied to each component of its input independently. For example, $x$ is first multiplied by a vector (matrix of a single column) $W_{i}^{0}$ and then each component goes through $\sigma$. The result is then multiplied by matrix $W_{ji}^{1}$ and again nonlinear $\sigma$ is applied. Process is repeated as many times as there are layers in a network\footnote{Although in applications vectors of thresholds are added before each nonlinear mapping. Here they are omitted for brevity.}. This one has four of them. Numbers of neurons in hidden layers set ranges for $i$, $j$ and $k$. Numerical parameters are represented by weights matrices:
\[
W_{i}^{0},W_{ji}^{1},W_{kj}^{2},W_{k}^{3}
\]
and they can be tuned by methods like gradient descent\cite{bryson1975applied,rumelhart1988learning}. Functions of this class are able to represent arbitrary mappings in various dimensions\cite{hornik1989multilayer,cybenko1989approximation,kuurkova1992kolmogorov,hornik1991approximation}, as well as their derivatives\cite{hornik1991approximation,cardaliaguet1992approximation}.
\section{Background}
%
%
%
%
Feedforward neural networks are able to solve partial differential
equations\cite{lagaris1998artificial,kumar2011multilayer,lagaris1997artificial} by weights minimization technique that uses the equation itself as a cost function, so, if it reaches small enough values, one can conclude that the equation holds within some error margin.
The input of such network is considered as a vector of independent variables
and the output as value of solution. All necessary derivatives of output
with respect to input and of cost function with respect to weights
can be calculated by the extended backpropagation procedure\cite{avrutskiy2017}. Including
boundary conditions into cost function itself usually does not produce
very accurate results\cite{shirvany2009multilayer}, so a process of function substitution
is required\cite{lagaris1998artificial}. For brevity, 2D case will be described, however, generalization
is straightforward. Consider a boundary value problem for partial
differential equation written for function $u(x,y)$:
\[U(x,y,u,u_{x},u_{y},...)=0\]
in a region $\Gamma$ with boundary $\partial\Gamma$ and condition:
\[\left.u\right|_{\partial\Gamma}=f\]
A new function $v$ is introduced by relation: 
\[
u(x,y)=v(x,y)\cdot\phi(x,y)+f
\]
where $\phi$ is known and carefully chosen to be smooth, vanish
on $\partial\Gamma$: 
\[\left.\phi\right|_{\partial\Gamma}=0\] 
to have reasonable normal derivative on the boundary:
\[
\left.\frac{\partial\phi}{\partial n}\right|_{\partial\Gamma}\sim1
\]
and to behave on $\Gamma$ ``as simply as possible''. The complete set of
requirements for such functions has not yet been formulated. After
writing the equation for $v(x,y)$, one can notice that boundary condition
turns into:
\[
\left.v\right|_{\partial\Gamma}<\infty
\] 
It seems that it's not necessary to account for that during the weights minimization since neural
networks do not converge to infinite valued functions if there is
a finite solution nearby. In applications $\phi$ is usually chosen
as the simplest analytical expression that vanishes on boundary and has
maximum value equal to 1, which is reached at a single point inside
region $\Gamma$. Condition $\frac{\partial\phi}{\partial n}\sim1$
is imposed to exclude situations when said derivative is too small.
Otherwise, $v$ near the boundary would have to be too large in order
to produce correct $\frac{\partial u}{\partial n}$, which is expected to be of the order of 1. In trivial cases,
like boundary problem inside a sphere of radius $1$, one can choose:
\[
\phi=1-r^{2}
\]
After substitution, the equation for $v(x,y)$ can be written as
\[V(x,y,v,v_{x},v_{y},...)=0\]
Minimization will use cost function $e=V^{2}$. A set of grid points
$(x_{i},y_{i})$, $i\in1...N$ inside the region $\Gamma$ is generated
as a matrix $X$ with two rows and $N$ columns. All derivatives that
are encountered in $V$ for $v$ like $\frac{\partial}{\partial x}$,$\frac{\partial}{\partial y}$
and so on have to be calculated for $X$. For example, $\frac{\partial}{\partial x}X$
is a matrix with the first row elements equal to 1 and the second row ones
equal to 0, vice versa for $\frac{\partial}{\partial y}X$.
All higher derivatives of $X$ are zero. Said set of matrices:\[
X,\frac{\partial}{\partial x}X,\frac{\partial}{\partial y}X,...\]
is propagated forward from one layer to another, and when the final
layer is reached, an output set is obtained: 
\[
v_{i},\frac{\partial}{\partial x}v_{i}\equiv v_{x}^{i},
\frac{\partial}{\partial y}v_{i}\equiv v_{y}^{i},...
\]
All of its members are matrices of size $1\times N$. Cost function is discretized as:
\[
E=\underset{x,y\in\Gamma}{\mu}(e)\simeq\frac{1}{N}{\sum_{i=1}^{N}}V^{2}(x_{i},y_{i},v_{i},v_{x}^{i},v_{y}^{i},...)
\]
Extended backward pass requires calculating derivatives of $E$ with
respect to each element of the output matrices: 
\[
\frac{\partial E}{\partial v_{i}},
\frac{\partial E}{\partial v_{x}^{i}},\frac{\partial E}{\partial v_{y}^{i}},...
\]
and these derivatives are gathered in matrices of the same size
$1\times N$. They are propagated backward in order to obtain gradient
of $E$ with respect to the weights of each layer. The whole procedure
is described in\cite{avrutskiy2017}. One can mention a few features of the neural network approach:
\begin{itemize}
\item Derivatives are calculated analytically, so numerical effects can
only be brought by rounding errors and/or incorrect discretization of the cost function's measure $\mu$, that can happen when grids have large spacing. 

\item The solution is obtained as a neural network, which is a smooth closed-form expression.
It exists naturally in the entire region rather than on finite set of points. 
This allows to modify the grid during the training as much as necessary without algorithm forgetting solution's values in points that were removed.

\item Parallelization is trivial: dense grids can be split into parts and processed by different computational nodes, which would only have to synchronize few megabytes
of weights data during each step regardless of the grid size.

\item Training is based on matrix multiplications and its careful implementation allows to achieve very high hardware efficiency on modern GPUs\cite{avrutskiy2017}.

\end{itemize}

\section{Method's extension}
Previously described approach is more or less standard for direct
solving of partial differential equations with neural networks, although
no papers prior to \cite{avrutskiy2017} formulated it in a form that was suitable for networks of arbitrary topology. In the same study the method was enhanced by adding extra terms to cost function. Namely, for equation:
\[
V(x,y,v,v_{x},...)=0
\]
one can write a trivial consequence: 
\[
\frac{\partial}{\partial x}V=0
\]
and include it into cost $e=V^{2}+V_{x}^{2}$. Applying operators
\[
\frac{\partial}{\partial x},
\frac{\partial}{\partial y},
\frac{\partial^{2}}{\partial x^{2}},
\frac{\partial^{2}}{\partial y^{2}}
\]
to $V$ and adding the corresponding terms to $e$ for
2D case allowed to increase precision and avoid overfitting to such
a degree that solving boundary value problem on a grid of 59 points
inside a unit circle produced $2\cdot10^{-5}$ accurate result for
the entire region, whereas training with no extra terms could only achieve accuracy of $2\cdot10^{-3}$\cite{avrutskiy2017overfit}. In this paper additional derivatives up to the fourth order are calculated along two directions $\xi$ and $\zeta$, which are random orthonormal vectors in space of independent variables. Since the cost
function now includes terms like $V_{\xi}^{2}$ and $V_{\zeta}^{2}$,
which depend on $v_{\xi}$ and $v_{\zeta}$ and so on, it is necessary to initialize matrices like $\frac{\partial}{\partial\xi}X$ and $\frac{\partial}{\partial\zeta}X$
at the input layer of the network. Using coordinates in $(x,y)$ basis: \[\xi=(a_{1},a_{2}),\zeta=(b_{1},b_{2})\] 
one can write:
\[
\frac{\partial}{\partial\xi}=a_{1}\frac{\partial}{\partial x}+a_{2}\frac{\partial}{\partial y}
\]
and similarly for $\frac{\partial}{\partial\zeta}$. Therefore, $\frac{\partial}{\partial\xi}X$
is a linear combination of $\frac{\partial}{\partial x}X$ and $\frac{\partial}{\partial y}X$
with coefficients $a_{1}$ and $a_{2}$, and so is $\frac{\partial}{\partial\zeta}X$,
but with $b_{1}$ and $b_{2}$. Higher order terms like $\frac{\partial^{2}}{\partial\xi^{2}}X$
are all zeros. Those extra matrices are used by forward pass to calculate
derivatives $v_{\xi}$, $v_{\zeta}$, $v_{\xi\xi}$ and others on the output
layer of the perceptron. 

Forward propagation works independently for every grid point $(x_{i},y_{i})$,
and each point with index $i$ is represented by $i$\textsuperscript{th} columns of all matrices initialized at the input layer. Noting that columns of $\frac{\partial}{\partial\xi}X$
and $\frac{\partial}{\partial\zeta}X$ are coordinates of vectors
$\xi$ and $\zeta$ in $x,y$ basis, one can conclude that different
points can have different and, more importantly, random directions
$\xi_{i}$ and $\zeta_{i}$. To implement that, one can generate $2N$
uniformly distributed unit vectors, split them into pairs, run Gram-Schmidt
process on each pair, and then write one vector to $\frac{\partial}{\partial\xi}X$
and another to $\frac{\partial}{\partial\zeta}X$. For 2D case $\xi_{i}$
and $\zeta_{i}$ form a basis in the input space that is randomly rotated
from point to point. For higher dimensions it is possible to include
more directional derivatives that will not be a linear combination
of $\frac{\partial}{\partial\xi}$ and $\frac{\partial}{\partial\zeta}$, but since $\xi$ and $\zeta$ are different in each point, all directions
of the input space are somewhat covered. Using one direction instead of
two was found to produce less accurate results, and using three directions
did not bring much improvement in higher dimensions.

In all cases training is split into three phases that use extra derivatives
up to fourth order. Cost functions are built using the following terms:
\[
V_{0}=V
\]
\[
V_{1}=V_{\xi}^{2}+V_{\zeta}^{2}
\]
\[
V_{2}=V_{\xi\xi}^{2}+V_{\zeta\zeta}^{2}
\]
\[
V_{3}=V_{\xi\xi\xi}^{2}+V_{\zeta\zeta\zeta}^{2}
\]
\[
V_{4}=V_{\xi\xi\xi\xi}^{2}+V_{\zeta\zeta\zeta\zeta}^{2}
\]
The first phase uses terms from 0 to 4. The second one uses terms from 0 to
3 and the last one from 0 to 2. Cost functions can be written
as:
\[e_{s}=\sum_{j=0}^{s}V_{j}\] 
for $s=4$, $3$ and $2$. They spawn a considerable amount of matrices that one should propagate. For example, 5D Poisson equation itself requires 2 derivatives for each dimension, 10 in total: 
\begin{equation}\label{operatori}
\frac{\partial}{\partial x_{j}},\frac{\partial^{2}}{\partial x_{j}^{2}},j\in\{1...5\}
\end{equation}
Then 8 additional operators are casted:
\begin{equation}
\label{esheoperatori}
\frac{\partial}{\partial\xi},\frac{\partial^{2}}{\partial\xi^{2}},\frac{\partial^{3}}{\partial\xi^{3}},\frac{\partial^{4}}{\partial\xi^{4}}
\end{equation}
\begin{equation}\label{iesheoperatori}
\frac{\partial}{\partial\zeta},\frac{\partial^{2}}{\partial\zeta^{2}},\frac{\partial^{3}}{\partial\zeta^{3}},\frac{\partial^{4}}{\partial\zeta^{4}}
\end{equation}
Results of \eqref{esheoperatori}, \eqref{iesheoperatori}, their combinations with \eqref{operatori} and function values themselves constitute to a total number of 99 different matrices. For $s=3$ it's 77 and for $s=2$ it's 55.
\subsection{Renormalization}
For certain functions derivatives of various orders have vastly
different magnitude. For example, $\sin10x$ has values of the order
of 1 and its $n$\textsuperscript{th} derivative with respect to $x$ is of the order of $10^{n}$. If solution
of a partial differential equation demonstrates similar properties,
backpropagation will mostly minimize high order terms without proper
attention to low order ones. Such cases require normalization. For
each grid point the largest derivative along additional direction
is spotted: 
\[
M_{\xi}=\max(v_{\xi},v_{x\xi},v_{xx\xi},v_{\xi\xi},v_{x\xi\xi},v_{xx\xi\xi},...)
\]
If $M_{\xi}>4$, vector $\xi$ is divided by $\sqrt[k]{M_{\xi}}$,
where $k$ is the order of the largest derivative with respect to $\xi$.
Similar procedure is done for $\zeta$. Renormalization requires one
forward pass and is usually done once after each portion of epochs $r_{int}$. Note that even if solution does not produce derivatives of various magnitude, 
neural network can still exhibit this behavior somewhere along the training.
\section{Results}
To have non trivial analytical solutions, against which one could verify
numerical results, formulas $u_{a}$ for those solutions were first picked
and then substituted into linear:
\[
\triangle u=g
\]
and nonlinear Poisson equation:
\[
\triangle u+u^{2}=h
\]
to find the corresponding source terms $g$ and $h$ that would produce
them. In all cases equations are solved inside $n$-dimensional ball of radius 1 with vanishing boundary condition:
\[\Gamma:r<1\]
\[\left.u\right|_{\partial\Gamma}=0\]
which is then simplified by function substitution:
\[
u=v\cdot(1-r^{2})
\]
After numerical solutions are obtained for $v$, functions $u$ are calculated using them and compared against analytical expressions $u_{a}$ on dense enough sets of random points distributed uniformly inside $\Gamma$. Maximum and median values of absolute error ${\varepsilon=|u-u_{a}|}$ are calculated.

Grids are comprised of two parts. Surface part contains equidistant points from $\partial\Gamma$ with angular distance $\theta$. Internal part is a regular grid, which is generated in three steps. At first, a Cartesian grid with spacing $\lambda$ inside $[-1,1]^{n}$ is created. Two random vectors in $\mathbb{R}^{n}$ are then chosen, and the grid is rotated from one to another. Finally, it is shifted along each direction by random value from interval $[-\frac{\lambda}{4},\frac{\lambda}{4}]$, and after that all points with $r>1$ are excluded.

Training is based on RProp\cite{riedmiller1993direct} procedure with parameters $\eta_{+}=1.2$, $\eta_{-}=0.5$.
Weights are forced to stay in $[-20,20]$ interval, no min/max bonds
for steps are imposed. Initial steps $\Delta_{0}$ are set to $2\cdot10^{-4}$, unless otherwise stated. Weights for neurons are initialized\cite{glorot2010understanding,he2015delving}
with random values from range $\pm2/\sqrt{s}$, where $s$ is the number of senders. Thresholds are initialized in range $\pm0.1$. All layers, but input and output, are nonlinear with activation function:
\[\sigma(x)=\frac{1}{1+\exp(-x)}\]

The procedure is implemented according to \cite{avrutskiy2017} with no intermediate load/saves and run on Google Cloud instance with Nvidia Tesla P100 using 32 bit precision. All training points are processed in one batch. During the training original equation's residual $U_{0}=V_{0}$ is monitored, and its root mean squared is calculated for grid points. Training phases are thoroughly described for 2D case and applied without changes to higher dimensions. After figuring out grid requirements and network topologies all solutions were obtained from the first try.

\subsection{Two dimensions}
Analytical solution is written as:
\[
u(x_{1},x_{2})=\frac{10}{17}(1-r^{2})(x_{1}+\sin x_{2}+x_{1}^{2}+x_{2}\cos x_{1})
\]
\[
r^{2}=x_{1}^{2}+x_{2}^{2}
\]
Ratio $\frac{10}{17}$ is introduced, so that $u_{max}-u_{min}\simeq1$.
Source term for linear equation is:
\begin{align*}
g=& -\frac{10}{17}\left(-x_{1}^{2}-x_{2}^{2}+1\right)\sin x_{2}-\\
& -\frac{40}{17}x_{1}\left(2x_{1}-x_{2}\sin x_{1}+1\right)+\\
& +\frac{10}{17}\left(-x_{1}^{2}-x_{2}^{2}+1\right)\left(2-x_{2}\cos x_{1}\right)-\\
& -\frac{40}{17}x_{2}\left(\cos x_{1}+\cos x_{2}\right)-\\
& -\frac{40}{17}\left(x_{1}^{2}+x_{1}+\sin x_{2}+x_{2}\cos x_{1}\right)
\end{align*}
and for nonlinear equation is:
\begin{align*}
h=& -\frac{10}{17}\left(-x_{1}^{2}-x_{2}^{2}+1\right)\sin x_{2}-\\
& -\frac{40}{17}x_{1}\left(2x_{1}-x_{2}\sin x_{1}+1\right)+\\
& +\frac{10}{17}\left(-x_{1}^{2}-x_{2}^{2}+1\right)\left(2-x_{2}\cos x_{1}\right)-\\
& -\frac{40}{17}x_{2}\left(\cos x_{1}+\cos x_{2}\right)-\\
& -\frac{40}{17}\left(x_{1}^{2}+x_{1}+\sin x_{2}+x_{2}\cos x_{1}\right)+\\
& +\frac{100}{289}\left(-x_{1}^{2}-x_{2}^{2}+1\right){}^{2}\Bigl(x_{1}^{2}+x_{1}+\sin x_{2}+\\
& +x_{2}\cos x_{1}\Bigr)^{2}
\end{align*}
In both cases new function $v$ is introduced:
\[
u(x_{1},x_{2})=v(x_{1},x_{2})\cdot(1-x_{1}^{2}-x_{2}^{2})
\]
Linear equation turns into:
\begin{equation}\label{linv}
(1-x_{1}^{2}-x_{2}^{2})\triangle v-4x_{1}\frac{\partial v}{\partial x_{1}}-4x_{2}\frac{\partial v}{\partial x_{2}}-4v=g
\end{equation}
and nonlinear into:
\begin{equation}\label{nelinv}
\begin{aligned}
& (1-x_{1}^{2}-x_{2}^{2})\triangle v -4x_{1}\frac{\partial v}{\partial x_{1}}-4x_{2}\frac{\partial v}{\partial x_{2}}-\\
& -(1-x_{1}^{2}-x_{2}^{2})^{2}v^{2}-4v =h
\end{aligned}
\end{equation}
Boundary conditions are now trivial:
\[
\left.v\right|_{\partial\Gamma}<\infty
\]
Neural network is a fully connected perceptron with the following layer structure:
\[
2,96,96,96,96,96,96,1
\]
\subsubsection{Linear 2D equation} 
\begin{figure}[!t]
	\centering
	\includegraphics[width=3.0in]{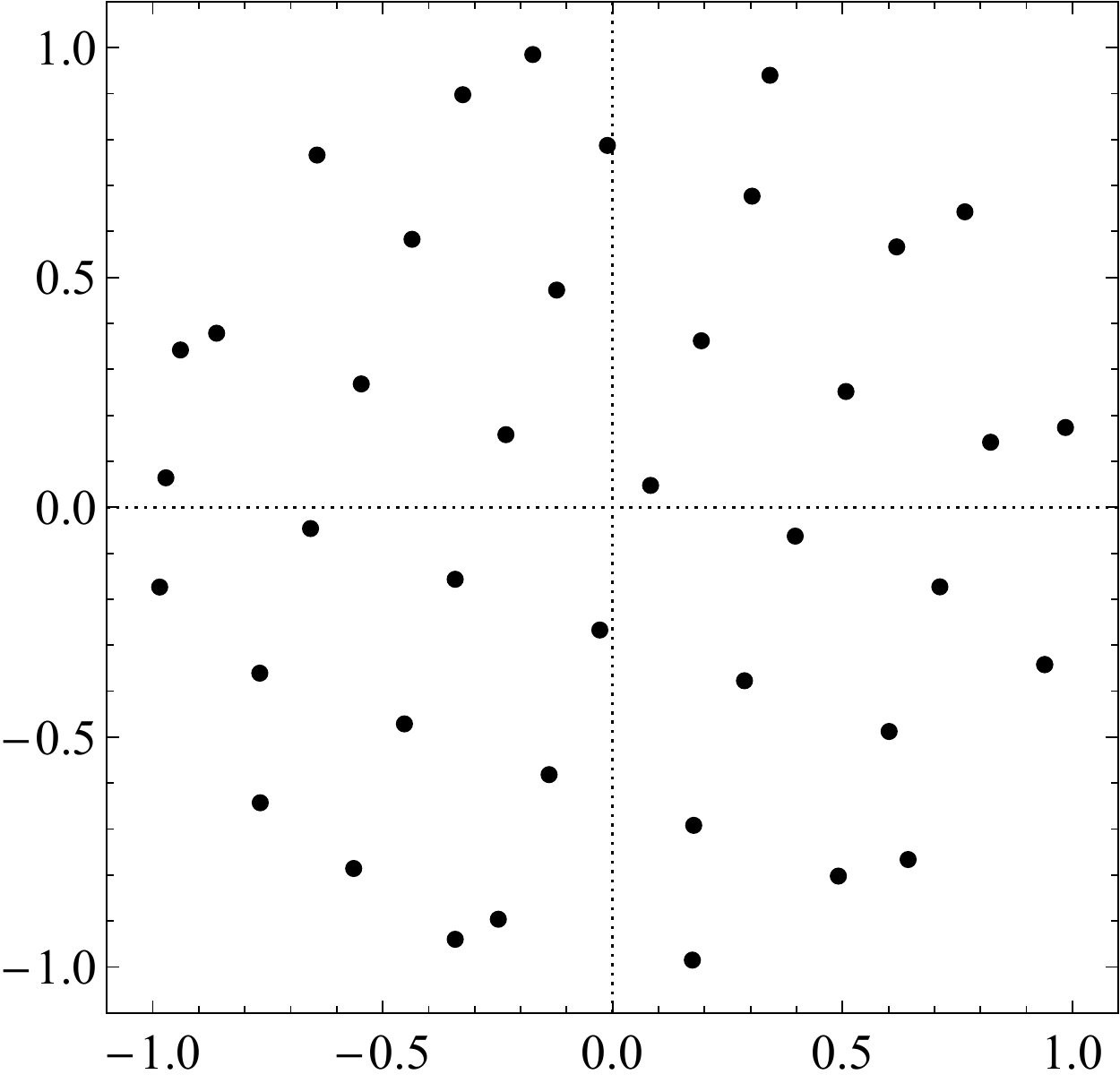}
	\caption{2D grid for linear Poisson equation. In higher dimensions the same spacing is used.}
	\label{fig:2dg}
\end{figure}

Surface grid contains 13 points of a circle with $\theta=\frac{\pi}{6}$. Internal grid has spacing $\lambda=\frac{1}{3}$ and contains 27 points, the total number is 40 (see Fig. \ref{fig:2dg}).

The first training phase uses $e_{4}$ and lasts for 2000 epochs with $r_{int}=200$. After the first 1000 epochs, steps $\Delta$ are reset back to $2\cdot10^{-4}$. RMS of equation's residual at the end: $\bar{V}_{0}=4\cdot10^{-4}$.

The second phase uses $e_{3}$, $\Delta_{0}=2\cdot10^{-5}$ and lasts for 2000 epochs with $r_{int}=200$. At the end of it $\bar{V}_{0}=1.3\cdot10^{-4}$.
 
The third phase uses $e_{2}$, $\Delta_{0}=2\cdot10^{-5}$ and lasts for 2000 epochs with $r_{int}=200$. The final RMS of the residual: ${\bar{V}_{0}=4\cdot10^{-5}}$.

$u(x_{1},x_{2})$ is obtained in 114 seconds on grid of 40 points and tested on 4000 points against the analytical solution with the following results:
\[
\varepsilon_{max}=3.5\cdot10^{-6},\varepsilon_{median}=6.8\cdot10^{-7}
\]

\subsubsection{Nonlinear 2D equation}
Surface grid contains 17 points of a circle with $\theta=\frac{\pi}{8}$. Internal grid has spacing $\lambda=\frac{1}{4}$ and contains 47 points, the total number is 64.

The first training phase uses $e_{4}$, lasts for 3000 epochs and is split into 5 intervals. The first one contains 160 epochs, $r_{int}=15$. The second has 340 epochs and $r_{int}=50$. The third is 500 epochs and $r_{int}=50$. The fourth is 1000, $r_{int}=50$ and the final one is another 1000 epochs with $r_{int}=100$. Steps $\Delta$ are reset back to $2\cdot10^{-4}$ at the end of each interval but the last one. At the end of this phase $\bar{V}_{0}=4\cdot10^{-2}$.

The second phase uses $e_{3}$, $\Delta_{0}=2\cdot10^{-5}$ and lasts for 2000 epochs with $r_{int}=200$. At the end of it $\bar{V}_{0}=1\cdot10^{-3}$.

The third phase uses $e_{2}$, $\Delta_{0}=2\cdot10^{-5}$ and lasts 2000 epochs with $r_{int}=200$. The final RMS of the residual $\bar{V}_{0}=1\cdot10^{-4}$.

$u(x_{1},x_{2})$ is obtained in 180 seconds on grid of 64 points and tested on 4000 points against the analytical solution with the following results:
\[
\varepsilon_{max}=8\cdot10^{-6},\varepsilon_{median}=2.4\cdot10^{-6}
\]

\subsection{Three dimensions}
Analytical solutions is written as:
\[
u(x_{1},x_{2},x_{3})=\frac{3}{5}(1-r^{2})(x_{1}+\sin x_{2}+x_{3}^{2}+x_{2}\cos x_{1})
\]
\[
r^{2}=x_{1}^{2}+x_{2}^{2}+x_{3}^{2}
\]
Substitution of function is implemented in a similar manner:
\[
u(x_{1},x_{2},x_{3})=v(x_{1},x_{2},x_{3})\cdot(1-r^{2})
\]
Equations and corresponding sources $g$ and $h$ are obtained by exactly the same procedure as in two-dimensional case. Results are similar and for brevity they are omitted. Neural network is a fully connected perceptron with the following layer structure:
\[
3,96,96,96,96,96,96,1
\]
\subsubsection{Linear 3D equation}
Surface grid contains 51 points of a sphere with $\theta=\frac{\pi}{6}$. Internal grid has spacing $\lambda=\frac{1}{3}$ and contains 108 points, the total number is 159.

The first phase is the same as phase 1 of 2D Linear case. At the end of it $\bar{V}_{0}=1.5\cdot10^{-3}$.

The second phase is the same as phase 2 of 2D Linear case and it ends with $\bar{V}_{0}=3\cdot10^{-4}$.

The third phase is the same as phase 3 of 2D Linear case and the final $\bar{V}_{0}=1.3\cdot10^{-4}$.

$u(x_{1},x_{2},x_{3})$ is obtained in 226 seconds on grid of 159 points and tested on 35000 points against the analytical solution with the following results:
\[
\varepsilon_{max}=6.5\cdot10^{-6},\varepsilon_{median}=1.4\cdot10^{-6}
\]

\subsubsection{Nonlinear 3D equation}
Surface grid contains 87 points of a sphere with $\theta=\frac{\pi}{8}$. Internal grid has spacing $\lambda=\frac{1}{4}$ and contains 265 points, the total number is 343.

The first phase is the same as phase 1 of 2D Nonlinear case and it leaves $\bar{V}_{0}=3.2\cdot10^{-2}$.

The second phase is the same as phase 2 of 2D Nonlinear case and it ends with $\bar{V}_{0}=6\cdot10^{-4}$.

The third phase is the same as phase 3 of 2D Nonlinear case and the final $\bar{V}_{0}=1.8\cdot10^{-4}$.

$u(x_{1},x_{2},x_{3})$ is obtained in 380 seconds on grid of 343 points and tested on 35000 points against the analytical solution with the following results:
\[
\varepsilon_{max}=8.2\cdot10^{-6},\varepsilon_{median}=1.9\cdot10^{-6}
\]

\subsection{Four dimensions}
Analytical solutions is written as:
\[
u(x_{1},x_{2},x_{3},x_{4})=\frac{7}{9}(1-r^{2})(x_{1}+\sin x_{2}+x_{3}^{2}+x_{4}\cos x_{4})
\]
\[
r^{2}=x_{1}^{2}+x_{2}^{2}+x_{3}^{2}+x_{4}^{2}
\]
Substitution is similar:
\[
u(x_{1},x_{2},x_{3},x_{4})=v(x_{1},x_{2},x_{3},x_{4})\cdot(1-r^{2})
\]
Neural network is a fully connected perceptron with the following layer structure:
\[
{4,148,148,148,148,148,148,1}
\]
\subsubsection{Linear 4D equation}
Surface grid contains 154 points of a 3-sphere with $\theta=\frac{\pi}{6}$. Internal grid has spacing $\lambda=\frac{1}{3}$ and contains 358 points, the total number is 512.

Phase 1 (see phase 1 of 2D Linear): $\bar{V}_{0}=1.5\cdot10^{-3}$.

Phase 2 (see phase 2 of 2D Linear): $\bar{V}_{0}=3.3\cdot10^{-4}$.

Phase 3 (see phase 3 of 2D Linear): $\bar{V}_{0}=1.7\cdot10^{-4}$.

$u(x_{1},x_{2},x_{3},x_{4})$ is obtained in 560 seconds on grid of 512 points and tested on 500,000 points against the analytical solution with the following results:
\[
\varepsilon_{max}=6.5\cdot10^{-6},\varepsilon_{median}=9.7\cdot10^{-7}
\]

\subsubsection{Nonlinear 4D equation}
Surface grid contains 357 points of a 3-sphere with $\theta=\frac{\pi}{8}$. Internal grid has spacing $\lambda=\frac{1}{4}$ and contains 1179 points, the total number is 1536.

Phase 1 (see phase 1 of 2D Nonlinear): $\bar{V}_{0}=5\cdot10^{-2}$.

Phase 2 (see phase 2 of 2D Nonlinear): $\bar{V}_{0}=1.2\cdot10^{-3}$.

Phase 3 (see phase 3 of 2D Nonlinear): $\bar{V}_{0}=2.5\cdot10^{-4}$.

$u(x_{1},x_{2},x_{3},x_{4})$ is obtained in 1160 seconds on grid of 1536 points and tested on 500,000 points against the analytical solution with the following results:
\[
\varepsilon_{max}=8.6\cdot10^{-6},\varepsilon_{median}=1.2\cdot10^{-6}
\]

\subsection{Five dimensions}
Analytical solutions is written as:
\[
u(x_{1},x_{2},x_{3},x_{4},x_{5})=\frac{7}{9}(1-r^{2})(x_{1}+\sin x_{2}+x_{3}^{2}+x_{4}\cos x_{5})
\]
\[
r^{2}=x_{1}^{2}+x_{2}^{2}+x_{3}^{2}+x_{4}^{2}+x_{5}^{2}
\]
Substitution is as usual:
\[
u(x_{1},x_{2},x_{3},x_{4},x_{5})=v(x_{1},x_{2},x_{3},x_{4},x_{5})\cdot(1-r^{2})
\]
Neural network is a fully connected perceptron with the following layer structure:
\[
5,160,160,160,160,160,160,1
\]

\subsubsection{Linear 5D equation}
Surface grid contains 399 points of a 4-sphere with $\theta=\frac{\pi}{6}$. Internal grid has spacing $\lambda=\frac{1}{3}$ and contains 1137 points, the total number is 1536.

Phase 1 (see phase 1 of 2D Linear): $\bar{V}_{0}=1.5\cdot10^{-3}$.

Phase 2 (see phase 2 of 2D Linear): $\bar{V}_{0}=3.2\cdot10^{-4}$.

Phase 3 (see phase 3 of 2D Linear): $\bar{V}_{0}=2.1\cdot10^{-4}$.

$u(x_{1},x_{2},x_{3},x_{4},x_{5})$ is obtained in 1280 seconds on grid of 1536 points and tested on 5,000,000 points against the analytical solution with the following results:
\[
\varepsilon_{max}=8.7\cdot10^{-6},\varepsilon_{median}=1.2\cdot10^{-6}
\]

\subsubsection{Nonlinear 5D equation}
Surface grid contains 1217 points of a 4-sphere with $\theta=\frac{\pi}{8}$. Internal grid has spacing $\lambda=\frac{1}{4}$ and contains 5311 points, the total number is 6528.

Phase 1 (see phase 1 of 2D Nonlinear): $\bar{V}_{0}=3.2\cdot10^{-2}$.

Phase 2 (see phase 2 of 2D Nonlinear): $\bar{V}_{0}=2\cdot10^{-3}$.

Phase 3 (see phase 3 of 2D Nonlinear): $\bar{V}_{0}=3.3\cdot10^{-4}$.

$u(x_{1},x_{2},x_{3},x_{4},x_{5})$ is obtained in 5000 seconds on grid of 6528 points and tested on 5,000,000 points against the analytical solution with the following results:
\[
\varepsilon_{max}=8.8\cdot10^{-6},\varepsilon_{median}=1.5\cdot10^{-6}
\]

\section{Catching up with Finite Differences}
Linear Poisson equations will be considered in this section. For estimation of time required to solve those with classical methods, a simple $2n+1$ point stencil in $n$ dimensions is considered. For example, in 2D case it states:
\begin{align*}
\triangle u &= \frac{1}{h^{2}}(u(x_{1}+h,x_{2})+u(x_{1}-h,x_{2})-2u(x_{1},x_{2}))+\\
& +\frac{1}{h^{2}}(u(x_{1},x_{2}+h)+u(x_{1},x_{2}-h)-2u(x_{1},x_{2}))+\\
& +\frac{h^{2}}{12}(\frac{\partial^{4}u}{\partial x_{1}^{4}}+\frac{\partial^{4}u}{\partial x_{2}^{4}})+O(h^{4})
\end{align*}
In 5D it gives the following discretization error:
\[
\epsilon=\frac{h^{2}}{12}(\frac{\partial^{4}u}{\partial x_{1}^{4}}+\frac{\partial^{4}u}{\partial x_{2}^{4}}+\frac{\partial^{4}u}{\partial x_{3}^{4}}+\frac{\partial^{4}u}{\partial x_{4}^{4}}+\frac{\partial^{4}u}{\partial x_{5}^{4}})+O(h^{4})
\]
For analytical solution $u_{a}(x_{1},x_{2},x_{3},x_{4},x_{5})$ it turns into:
\begin{align*}
\epsilon= & \frac{7}{108}h^{2}\Bigl(-24-8x_{4}x_{5}\sin x_{5}+8x_{2}\cos x_{2}-\\
& -\left(x_{1}^{2}+x_{2}^{2}+x_{3}^{2}+x_{4}^{2}+x_{5}^{2}-13\right)\cdot\\
& \cdot\left(\sin x_{2}+x_{4}\cos x_{5}\right)\Bigr)+O(h^{4})
\end{align*}

For estimation this whole expression can be reduced to the first term:
\begin{equation}\label{epsilonocenka}
\epsilon\sim \frac{7\cdot24}{108}h^{2}\simeq 1.6\cdot h^{2}
\end{equation}
Now instead of original equation:
\[
\triangle u=h
\]
discretization results in solving:
\[
\triangle\widetilde{u}=h-\epsilon
\]
The difference $\Psi$ between analytical and discrete solutions can be obtained using the following equation:
\begin{equation}\label{raznica}
\triangle(u-\widetilde{u})=\triangle\Psi=\epsilon
\end{equation}
with vanishing boundary condition:
\[
\left.\Psi\right|_{\partial\Gamma}=0
\]
For constant $\epsilon$ it can be readily solved:
\[
\Psi=-\frac{\epsilon}{10}(1-x_{1}^{2}-x_{2}^{2}-x_{3}^{2}-x_{4}^{2}-x_{5}^{2})
\]
which gives:
\begin{equation}\label{ocenka}
\max|u-\widetilde{u}|\sim\frac{|\epsilon|}{10}
\end{equation}

Note that for other dimensions instead of $10$ one should put $2n$. This result is confirmed for low dimensions by solving \eqref{raznica} numerically with no terms of $\epsilon$ omitted. 

One can argue that since neural network is solving another equation - the one that is written for $v$, its discretization error $\epsilon_{v}$ is different. Closer look reveals that terms of $\epsilon_{v}$ related to Laplacian indeed become $\sim18$ times smaller, however the most part of the error now comes from the first order derivatives (see \eqref{linv}, \eqref{nelinv}):
\[
4x_{i}\frac{\partial v}{\partial x_{i}}
\]
Leading terms of $\epsilon_{v}$ for those:
\[
x_{i}\frac{4h^{2}}{6}\frac{\partial^{3}v}{\partial x_{i}^{3}}\sim 0.5 h^{2}
\]
The later equivalence is due to the third derivatives of the analytical solution:
\[
v_{a}=\frac{7}{9}(x_{1}+\sin x_{2}+x_{3}^{2}+x_{4}\cos x_{4})
\] being of the order of $7/9$. To make sure that the ratio of 3 between $\epsilon_{v}$ and $\epsilon$ is not a significant advantage for a neural network, the term $x^{2}_{3}$ in $v_{a}$ was replaced with $x^{3}_{3}/2$. This increases discretization error for $v$ by three times. After neural network solution was recalculated, the effect of that on precision turned out to be subtle and was fully compensated by setting ${r_{int}=20}$ for the first 1000 epochs of phase 1 and extending said phase for another 500 epochs. This increased total running time by 10\%. However, all those changes also produced even better final result: \[\varepsilon_{max}=6.5\cdot10^{-6},\varepsilon_{median}=1.13\cdot10^{-6}\]

Using established estimations for $\epsilon$ and setting maximum allowed error to $\delta=10^{-5}$, the grid spacing for the second order finite difference scheme can be written as:
\[
\sqrt{\frac{10\delta}{1.6}}\sim0.008
\]
It results in $125^{5}\simeq3\cdot10^{10}$ points for each unit of 5D volume. Solution is obtained in the region $\Gamma$, which is a ball of radius 1. Its volume is:
\[
\frac{8\pi^{2}}{15}\simeq5.26
\]
Therefore, to cover internal part of $\Gamma$ $1.6\cdot10^{11}$ points are needed. The lower bond for the size of the external part can be calculated using similar considerations: 4D surface of 5D sphere has a measure of:
\[
\frac{8\pi^{2}}{3}\simeq26.3
\]
And it should at least be uniformly covered by points with the same spacing, which gives:
\[
125^{4}\cdot26.3\simeq6.4\cdot10^{9}
\]
This value is about $5\%$ of the internal one and is going to be omitted.

After discretization a linear system is to be solved. Its number of variables is equal to that of the grid points. Matrix for such system is $2n+1=11$ diagonal. Solving time will be extrapolated using the data from H. Liu et al.\cite{liu2015accelerating}. They investigated performance of parallel AMG solvers run on Nvidia Tesla C2070 for various matrices and different implementations. Apart from relatively fast solution times\cite{markov2013high,liu2014accelerating,krotkiewski2013efficient} they use similar hardware, which eases the task of extrapolation. Tesla C2070 achieves 515 GFLOPS as it was used in double precision mode. In this paper Tesla P100 is operating with single precision for which it delivers 9340 GFLOPS. To make some room for estimation error one can consider that AMG solvers can run on single precision just as well. Therefore, the total ratio between theoretical computational powers is 18. The most similar matrix is $P3D7P\textunderscore100$. It was created by similar 7 point stencil for 3D Poisson equation on $100^{3}$ grid. The least processing time mentioned for this matrix is $0.34$ seconds. Using linear extrapolation, one can write an approximate running time for solving 5D linear Poisson equation:
\[
t_{5D}=0.34\text{sec}\cdot\frac{1}{18}\frac{1.6\cdot10^{11}}{100^{3}}\simeq3000\text{sec}
\]
Estimation does not account for a fact that $P3D7P\textunderscore100$ is 7 diagonal, and the matrix to be solved is 11 diagonal. Similarly, for lower dimensions one can find:
\[
t_{4D}\simeq23\text{sec}
\]
\[
t_{3D}\simeq0.15\text{sec}
\]
\[
t_{2D}\simeq0.001\text{sec}
\]
\begin{figure}[!t]
	\centering
	\includegraphics[width=3.5in]{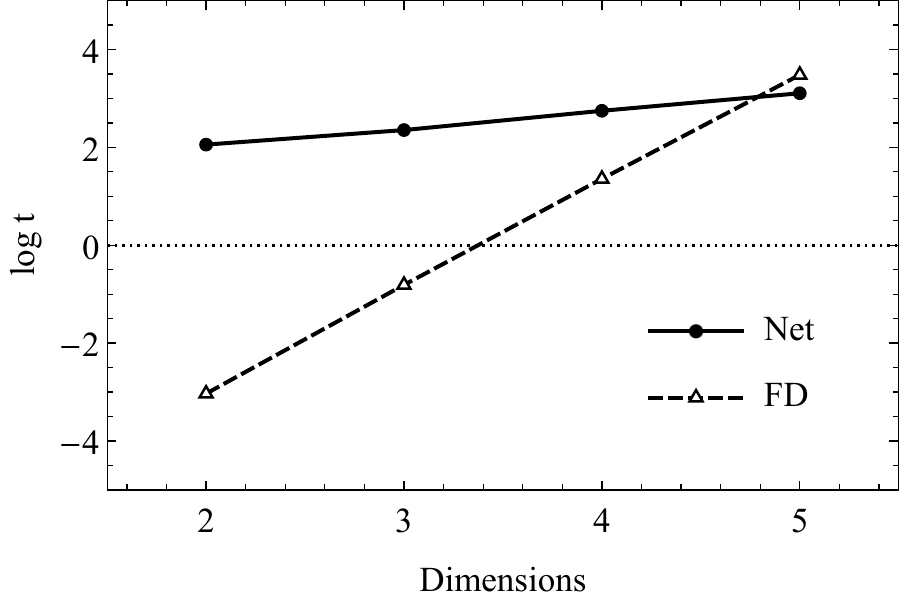}
	\caption{Base 10 logarithms of solving time using neural network and projected running time for the second order finite difference scheme.}
	\label{logt}
\end{figure}
Plots of solving time against the number of dimensions can be seen on Fig. \ref{logt}.

\section{Conclusion}
Neural network method was applied to boundary value problems for linear and nonlinear Poisson equations inside 2, 3, 4 and 5 dimensional unit balls. It's based on direct approximation of functions by multilayer perceptrons: independent variables are fed into input layer, and solution's values are obtained at the output. Procedure was improved by considering, in addition to equation itself, it's trivial consequences, obtained by differentiation. Prior to enhancements the method was mostly fringe and used as a proof of concept. With the additional cost terms and hardware efficient implementation it was able to compete with a second order finite difference scheme in solving 5D boundary value problem. For 4D cases and lower, classical method was far more effective. In higher dimensions neural networks managed to get ahead due to their ability to operate on grids with spacing as low as $1/3$ provided solution's derivatives up to 6th were of the order of 1. Even though it was compared against the simplest second order scheme, one can imagine some hard times trying to apply high order stencils to a grid that has no more than 6 points in each direction. Therefore, regardless of the order, the minimal number of points per dimension will probably always stay lower for neural networks so there is always a high dimensional task, for which finite differences are slower. One can also mention a decrease in memory complexity: if obtained on a grid, 5D solution would utilize at least 640GB of memory whereas neural network required about 2GB during the training and 640KB to store the result. This gap is mostly due to ineffective usage of resources by classical methods. For example, values of $125^5$ grid can easily describe 125 completely different solutions of 4D equations, whereas training a 5D neural network to represent such a mess would probably not end very well. On the other hand for functions that do not change too fast almost any grid representation is a waste of memory since a set of independent values has too many degrees of freedom. From this point of view neural networks can provide much more efficient representation, which matters for higher dimensions.


%
\appendices

\section*{Acknowledgment}
Author is extremely grateful to his scientific advisor E.A.~Dorotheyev for help and enormous amount of support during this research. Special thanks are due to Y.N.~Sviridenko, A.M.~Gaifullin, I.A.~Avrutskaya and I.V.~Avrutskiy without whom this work would be impossible.

\end{document}